# LightHCG: a Lightweight yet powerful HSIC Disentanglement based Causal Glaucoma Detection Model framework

Daeyoung, Kim

*Abstract*— As a representative optic degenerative condition, glaucoma has been a threat to millions due to its irreversibility and severe impact on human vision fields. Mainly characterized by dimmed and blurred visions, or peripheral vision loss, glaucoma is well known to occur due to damages in the optic nerve from increased intraocular pressure (IOP) or neovascularization within the retina. Traditionally, most glaucoma related works and clinical diagnosis focused on detecting these damages in the optic nerve by using patient data from perimetry tests, optic papilla inspections and tonometer-based IOP measurements. Recently, with advancements in computer vision AI models, such as VGG16 or Vision Transformers (ViT), AI-automatized glaucoma detection and optic cup segmentation based on retinal fundus images or OCT recently exhibited significant performance in aiding conventional diagnosis with high performance. However, current AI-driven glaucoma detection approaches still have significant room for improvement in terms of reliability, excessive parameter usage, possibility of spurious correlation within detection, and limitations in applications to intervention analysis or clinical simulations. Thus, this research introduced a novel causal representation driven glaucoma detection model: LightHCG, an extremely lightweight Convolutional VAE-based latent glaucoma representation model that can consider the true causality among glaucoma-related physical factors within the optic nerve region. Using HSIC-based latent space disentanglement and Graph Autoencoder based unsupervised causal representation learning, LightHCG not only exhibits higher performance in classifying glaucoma with 93~99% less weights, but also enhances the possibility of AI-driven intervention analysis, compared to existing advanced vision models such as InceptionV3, MobileNetV2 or VGG16.

*Key Words* — Glaucoma, Deep learning, HSIC, Causal representation learning, Computer vision, Retinal fundus image

I. INTRODUCTION

Glaucoma is a representative optic degenerative condition that damages the optic nerve or retinal ganglion cells with no specific symptoms in the early stages of progression, increasing the possibility of blindness within affected patients [1,2]. Primarily caused by increased intraocular pressure or neovascularization, symptoms of glaucoma are mainly characterized by deterioration of vision qualities, such as blurred and dimmed images [1], or peripheral vision loss within patients. In terms of the number of patients, approximately 76 million (2020) worldwide was estimated to suffer from glaucoma. Furthermore, the total number of patients who suffer from glaucoma is predicted to increase to 112 million by 2040 [3]. Being a significant threat to humanity due to its prevalence and the irreversibility of damaged optic nerves, most research has focused on detecting and preventing acceleration of glaucoma progression with diverse measures. Traditional approaches to detect glaucoma mainly implemented tonometer-based intraocular pressure(IOP) evaluation, optic papilla inspections from retinal fundus image analysis, or automated perimetry driven visual field tests such as Goldmann perimetry tests, which are all still one of the most precise and significant measures in ophthalmology [4,5]. However, with extreme advancements in computer vision AI models and influx of multimodal patient data, AI-based glaucoma detection, especially deep learning based glaucoma detection and segmentation exhibited immense progress which can aid ophthalmologist to diagnose glaucoma with significantly less resources. Specifically, transfer learning based approaches based on advanced vision models, such as EfficientNet, ViT, or ResNet, exhibited extremely competitive f1-scores or sensitivity scores within vision field/retinal fundus based glaucoma classification tasks [6,7]. Having significant success in accurately classifying status of glaucoma, computer vision models have now become tools that can provide guidance to clinicians in accurately detecting patients at risk of getting glaucoma within early stages.

However, current AI-driven glaucoma detection models tend to have high complexity in both parameter space and network structures, leading to high cost issues and reliability issues due to its nature as a black box. Though attention mechanisms and XAI techniques such as GradCAM partially provide insights regarding how the mechanism of a certain model is constructed, it is still a difficult task to comprehend the overall diagnosis process of current AI-based glaucoma classification within given approaches. Furthermore, as current vision models mostly focus on prediction performance itself, AI-based glaucoma detection solely relies on correlation within data, not the true causal or physical mechanisms of glaucoma, which in turn can lead to spurious correlations defining core features of the model. This can lead to less reliable decisions or inconsistent results when expanded to real-world medical field problems such as medical treatments or intervention simulations. Thus, to solve these potential issues and to provide an advanced framework for much more reliable AI-aided medical decisions, this research suggests a lightweight causal representation extraction model: LightHCG that incorporates causal representation learning and information based disentanglement to achieve robust, yet efficient representations of glaucoma within latent space from shallow VAEs. By considering the underlying causal physical mechanism of glaucoma progression via disentanglement within latent space, the proposed model and its framework not only enhance performance of AI-driven glaucoma diagnosis with extremely less parameters, but also expands the ability of MRI-based vision models from prediction to intervention effect simulations in terms of treatment.

## II. PRELIMINARIES

### A. Related Works

With recent advancements in computer vision models, active applications of advanced vision AI within glaucoma severity detection and segmentations can be found in various research. For example, in [6], based on Humphrey 24-2 SITA visual field test images and its pattern deviation plots from the UNSW Optometry Clinic and CFEH, pre-trained VGG16 and ResNet18 models were fitted to automatically classify glaucoma status (normal, early, moderate, advanced), which led to 83.92% and 87.77% 5-fold cross validation F1-scores when no image-augmentation was applied. Meanwhile, in [7], various transformer models, such as ViT, Swin Transformers, CrossViT and CaiT, were implemented to analyze retinal fundus images acquired from merging six datasets: LAG, ORIGA, ODIR-5K, HRF, REFUGE, DRISHTI-GS1. Under transfer learning based on ImageNet datasets, results show that sole use of Swin Transformers in merged retinal fundus test data succeeded in achieving the highest sensitivity of 92.57%, whereas use of CaiT succeeded in achieving the highest accuracy of 94.5%. Apart from ViT or VGG models, MobileNetV1 or MobileNetV2 models have also been implemented for glaucoma fundus classification under transfer learning [8]. In [8], with only 20 epochs of training under normal+glaucoma retina images, MobileNetV1 achieved 98% training accuracy and 72% testing accuracy, whereas MobileNetV2 achieved 96% training accuracy and 82% testing accuracy when global pooling and three FC layers were attached to baseline MobileNet architectures. Meanwhile, in [9], attempts that implement CNNs as feature encoders and traditional ML architectures as classifiers were tested under the ACRIMA retinal fundus dataset. Results show that among various combinations, CNN(5 convolutional layers and 5 consecutive max-pooling)+AdaBoost hybrid model succeeded in achieving the highest classification performance with 93.75% F1-score, 92.8% AUC and 92.96% accuracy, which was found to outperform classification accuracies from existing advanced CNN architectures, such as VGG16, ResNet50 or DenseNet121.

OCT based automatic glaucoma detections were also computed in several studies. For example, [10] introduced a ResNet-based 3D deep learning model which was applied in analyzing 4877 Spectral-domain OCT volumes from the Chinese University of Hong Kong Eye Centre and the Hong Kong Eye-Hospital to improve automatic glaucoma/non-glaucoma patient detection, achieving higher AUROC (96.9%) and validation accuracy (91%) scores when compared to 2D based deep learning systems and human glaucoma specialists. Meanwhile, [11] implemented an attention guided deep CNN model on 3D OCT images and corresponding VF test data for advanced glaucoma classification. Here, total loss of the proposed 3D CNN was defined as the total sum of (i) loss from using original 3D-cube OCT images for CNN inputs and glaucoma status as output, (ii) loss from using Grad-CAM heatmap-based occluded cubes as input and glaucoma status as output, and the (iii) loss from using attention-cropped cubes as input and glaucoma status as outputs. Results show that average predictions from pathway (i) to (iii) exhibits 93.8% AUC, 91.07% accuracy and 94.9% f1-scores.

On the other hand, not only classification tasks, but also automatic segmentation based on AI were probed to enhance glaucoma analysis. As a representative work, [12] introduced EARDS: EfficientNet and Attention-based Residual Depth-wise Separable Convolution which is an automated segmentation network for both the optic disk and cup. Adding an EfficientNet-b0 based encoder to the U-Net, attaching Attention Gate(AG) modules into original skip connections towards the decoder for implicit ROI learning, and implementing residual depth-wise separable convolution blocks within the decoder step, proposed framework of [12] achieved substantially significant dice coefficient scores of 97.41% and 95.49% from Drishti-GS dataset and REFUGE datasets.

Though success in automatic detection of glaucoma status from advanced vision models were sufficiently found, it was found that attempts to reduce the complexity of prevalent models such as Vision Transformers(ViT) or EfficientNet while preserving classification performance, or attempts to apply causality to reduce reliance on spurious correlations were found to be almost absent regarding related works. This research, therefore, can be defined as a crucial turning point in enhancing automatic glaucoma detection in the aspect of both efficiency and expandability(versatility), regarding real world clinical applications in glaucoma treatments.

### B. Physical mechanism of Glaucoma

Though there are diverse discussions regarding the primary factor or specific pathologies of glaucoma, most works agree on the fact that increased pressure within the retina contributes as a significant causal factor to the status of glaucoma. As pressure within the eye substantially increases due to factors such as abnormal fluid drainage in the anterior chamber, stress is accumulated to the lamina cribrosa, which then hinders axoplasmic flow [13,14]. This impaired transportation then damages the optic nerve, which is characterized by enlarged optic cups and neuroretinal rim thinning [15,16,17,18]. Specifically, decrease in the neuroretinal rim tissue directly affects limit of optic cup, which in turn affects the degree of exposure of the lamina [19]. (These optic nerve damages in rim and enlarged cups are also interpreted as 'retinal ganglion cell(RGC) death' that leads to distortions in visual fields). Meanwhile, abnormal retinal blood vessels from retinal pressure or neovascularization were also found to damage vision fields in terms of glaucoma [20,21].

Thus, the underlying mechanism of human glaucoma progression can be summarized as a causal graph in Fig 1. As pathologies regarding non-IOP related factors, especially neovascularization, are actively being probed recently, this research focused on direct physical factors emphasized in red (Fig 1). Using the causal link between optic cup, neuroretinal rim, vessel abnormality, and glaucoma as the ground truth causal graph, validity of results from causal representation learning within the LightHCG was thoroughly checked.

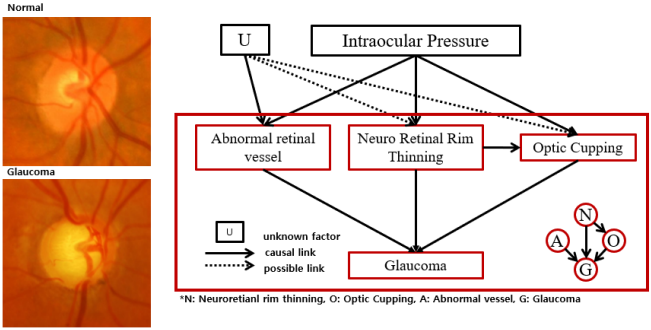

**Fig 1. Causal Mechanisms of Glaucoma (simplified)**

III. METHODOLOGY

A. Data

For comparison with existing works, this research implemented the ACRIMA dataset [22, 23], one of the most actively implemented public retinal fundus datasets in previous works. Consisted of 705 fundus images from the ACRIMA project(TIN2013-46751-R), the ACRIMA dataset provides 309 normal retinal images and 396 glaucomatous retinal images taken with Topcon TRC retinal camera and IMAGEnet capture systems. Labelling within ACRIMA images regarding existence of glaucoma was computed by two ophthalmology experts with 8 years of experience. Data can be accessed through Figshare or Kaggle [22,24]. Specifically, this research imported ACRIMA data from Kaggle, which provides preprocessed ACRIMA dataset images (cropping around the optic disc implemented). For robust model construction and representation extraction, ACRIMA dataset was randomly split into train : test images with 50:50 ratio. Here, training set was implemented for LightHCG latent space optimization and downstream model fitting, whereas test images were implemented to evaluate glaucoma prediction performance in downstream tasks. In terms of image size, all ACRIMA retinal images were resized to 224 x 224 format for analysis.

B. Proposed Framework of LightHCG

LightHCG implements a two-stage latent representation learning framework: (i) Hilbert Schmidt Independence Criterion(HSIC) based information disentanglement and a (ii) Graph Autoencoder(GAE) based causal representation learning under shallow Convolution Variational Autoencoder(VAE) architectures. Specifically, within the Convolutional VAE architecture, the primary latent space Z was split into two latent sub-spaces: $Z_1$ and $Z_2$, which were assumed to differentiate glaucoma-related and non-related features within the extracted latent space. The former space, $Z_1$ was assumed to incorporate non-glaucomatous features within retinal fundus images, whereas the latter ($Z_2$) was assumed as a sufficient representational space for glaucoma.

To further enhance the utility of glaucomatous latent sub-space, $Z_2$ in the aspect of clinical intervention analysis and robustness, causal representation learning was additionally implemented through the application of continuous neural causal discovery algorithms when training encoder structures directly linked to the formation of $Z_2$, thereby leading each latent variable in $Z_2$ to correspond with key causal factors dealt in Fig 1. By extracting only glaucoma related latent information and finding representations that can quantify three physical factors: neuroretinal rim thinning, optic cupping, and neovascularization with only shallow VAE structures, LightHCG not only aims to achieve a more reliable causal representation from retinal fundus images for both downstream tasks and intervention analysis, but also aims to extremely reduce the cost or burden of existing vision models, such as ViT or InceptionV3, in the aspect of parameter space. Specific concepts of individual components within LightHCG can be summarized as follows.

*(1) Graph Autoencoder-based Representation Learning under Convolutional VAE*

LightHCG implements a Convolutional VAE architecture to extract latent features from retinal images, while attaching a Graph Autoencoder-based causal discovery mechanism to latent sub-space $Z_2$ (Fig 2). Convolutional VAE, or CVAE is a CNN based Variational Autoencoder framework designed for image reconstruction or generation in diverse tasks. Similar to the vanilla VAE for tabular data analysis, CVAE aims to train a low-dimensional bottleneck latent space to incorporate or extract highly relevant features that can sufficiently represent input data, leading to the objective of extracting latent features from images under valid image reconstruction. Based on Evidence Lower Bound (ELBO) approaches [25], definition of loss within CVAE can be summarized as (1) and (2). In this research, as image data pixels are normalized to numerical values within 0 to 1, Bernoulli distributions were assumed for log likelihoods in computing negative ELBO, which converts the original CVAE reconstruction loss in (1) into a binary cross entropy loss in (2). Under the assumption of $q_\phi(z|x) \sim N(z, I)$ and $p(z) \sim N(0, I)$, regularization terms were reduced to the form of $0.5 \sum z_{ij}^2$, which leads to the final loss form of (2).

$$-\log(p(x)) \geq -\int \log\left(\frac{p(x,z)}{q_\phi(z|x)}\right) q_\phi(z|x) dz = -ELBO(\phi)$$
$$= -E_{q_\phi}(\log(p(x|z))) + KL(q_\phi(z|x)||p(z))$$
$$\simeq -\log \prod_{j=1}^{D} p(x^{(j)}|z^l) + KL(q_\phi(z|x)||p(z)), (L=1) \quad (1)$$

$$\sum_{j}^{D} \log p_{ij}^{x_{ij}} (1 - p_{ij})^{1-x_{ij}} + \frac{1}{2}\sum z_{ij}^2,$$
$$(q_\phi(z|x) \sim N(z,I), p(z) \sim N(0,I)) \quad (2)$$

Regarding the extracted latent sub-space $Z_2$ from CVAE, this research used Graph Autoencoder(GAE)-based causal discovery mechanisms from [26] to achieve valid causal representations of glaucoma. Similar to other neural continuous DAG structure learning algorithms such as NOTEARS [27] or DAG-GNN [28], Graph Autoencoder within causal discovery is based on the assumption of J, Pearl's Structural Causal Model (SCM) [29,30]. Applying Structural Equation Model (SEM) frameworks from Economics to causality, J, Pearl assumed that causal links within variables can be expressed through the linear equation of (3). Here, $A$ is considered as the adjacency matrix from underlying DAGs, whereas $\varepsilon$ is considered as a noise variable within causal links. In neural causal structure learning algorithms, elements in matrix $A$ are interpreted as trainable weights in neural networks, thereby expanding conventional discrete score-based DAG algorithms such as Hill Climbing (HC) to continuous parameter spaces. Converting DAG conditions to numerical constraints by setting $H(A)$ in equation (4) as 0, recent neural DAG search algorithms were found to succeed in retrieving diverse ground truth causalities from observational data.

Based on the success of neural causal discovery algorithms, this research attached Graph Autoencoder based DAG search to CVAE's second latent sub-space: $Z_2$. One significant variation within the proposed framework is that, unlike the original GAE of equation (5), this research assumed individual neural networks for all causal factors as in equation (6), so that the flexibility of generalized additive modelling frameworks in statistics can be incorporated in finding complex latent causal connections within components in retinal fundus images.

$$Z = A^T Z + \varepsilon, \quad \varepsilon \sim iid\ N(0, \Sigma), \quad \mathcal{G} = <A, Z> \quad (3)$$
$$H(A) = tr(e^{A \odot A}) - d, A \in \mathbb{R}^{d \times d} \quad (4)$$

$$Z = f_1\left(A^T f_2(Z)\right) + \varepsilon, \quad f_{1(2)} \approx Neural\ Network_{1(2)} \quad (5)$$
$$for\ \forall i^{th}\ variable\ Z_i, \quad \hat{Z}_i \simeq f_{2,i}\left(A_i^T f_{1,i}(Z)\right), f_{1(2),i}: NN \quad (6)$$

To incorporate DAG constraint of (4) within latent causal representation extractions, this research also implemented the augmented lagrangian methods introduced in [26] and [27]. As illustrated in (7) and (8), constraint: $h(A)=0$ was incorporated to the primary MSE reconstruction loss of Z by the lagrangian multiplier: $\alpha$, with an additional quadratic penalty term under $\rho > 0$ [31]. Here, parameters: $\alpha$ and $\rho$ were updated iteratively in each epoch apart from back propagations of gradients in neural networks.

Another significant variation within the proposed causal GAE framework is that corresponding glaucoma status variables were concatenated to latent vectors of $Z_2$ when GAE reconstructions were processed. That is, glaucoma status was weakly incorporated when finding latent causality within retinal fundus images to enable causal representations in $Z_2$ to correspond with direct physical factors dealt in Fig 1 and Section II. Preliminaries. Based on these previous settings, the overall LightHCG structure can be visualized as in Fig 2.

$$\min_{A, \theta_1, \theta_2} \mathcal{L}_{GAE} = \min_{A, \theta_1, \theta_2} \frac{1}{n} \sum_{i=1}^{n} \|Z^{(i)} - \hat{Z}^{(i)}\|^2 + \lambda \|A\|_1 + \alpha h(A)$$
$$+ \frac{\rho}{2}|h(A)|^2, \text{ where } h(A) = tr(e^{A \odot A}) - d \quad (7)$$

$$\alpha^{(k+1)} = \alpha^{(k)} + \rho^{(k)} h(A^{(k+1)})$$
$$\rho^{(k+1)} = \begin{cases} \beta \rho^{(k)}, (if\ |h(A^{(k+1)})| \geq \gamma |h(A^{(k)})|) \\ \rho^{(k)}, (o.w.), (\beta > 1, \gamma < 1) \end{cases} \quad (8)$$

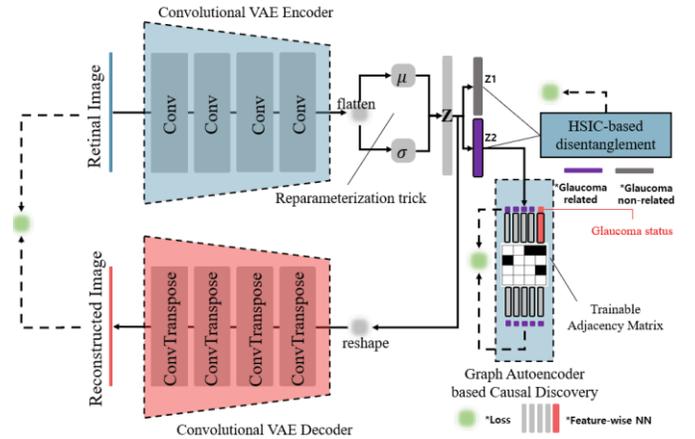

Fig 2. Visualization of the overall LightHCG framework

*(2) HSIC-based Information Disentanglement*

Meanwhile, splitting latent space from the convolutional variational autoencoder was computed based on the viewpoint of Information Theory. The intuition behind HSIC usage roots from the fact that the analyzed retinal image not only contains glaucoma related features, but also contains non-glaucomatous optic features such as basic shapes of the optic disk, camera-induced visual features, or pigments and vessel distribution in the outer part of the optic disc area. Thus, to extract robust causal glaucoma-related representations in $Z_2$, or in other words, for $Z_2$ to become a sufficient statistic of glaucoma status in the aspect of Data Processing Inequality(DPI) dealt in Information Bottleneck approaches, it becomes crucial to detect irrelevant retinal fundus information through robust metrics. This research, therefore, used Hilbert Schmidt Independence Criterion(HSIC) [32,33] as an alternative measure to Mutual Information(MI) to detect relevant/irrelevant fundus information and optimize representations within $Z_1$ and $Z_2$. HSIC is a measure that returns representations of statistical independencies between random variables (or vectors of variables) by computing the squared Hilbert-Schmidt (HS) norm from the associated cross-covariance operator between distributions [33]. Specifically, based on the theory of reproducing kernel Hilbert space(RKHS), basic definition of HSIC can be represented into equation (9) with cross-covariance defined as (10).

$$HSIC(p_{xy}, \mathcal{F}, \mathcal{G}) = \|C_{xy}\|_{HS}^2$$
$$\text{for } \forall x \in \mathcal{X}, \phi(x) \in \mathcal{F} \text{ exists s.t. } \langle \phi(x), \phi(x') \rangle_{\mathcal{F}} = k(x, x')$$
$$\text{for } \forall y \in \mathcal{Y}, \psi(y) \in \mathcal{G} \text{ exists s.t. } \langle \psi(y), \psi(y') \rangle_{\mathcal{G}} = l(y, y')$$
$$\mathcal{F}, \mathcal{G}: \text{both separable RKHS}, \quad k, l: \text{kernel} \quad (9)$$

$$\text{Cross Cov Operator}: C_{xy} \equiv E_{x,y}[(\phi(x) - \mu_x)(\psi(y) - \mu_y)]$$
$$= E_{x,y}[\phi(x) \otimes \psi(y)] - \mu_x \otimes \mu_y,; (C_{xy}: \mathcal{G} \to \mathcal{F}) \quad (10)$$

For computational purposes, definition of (9) can be converted into a kernel-based form of (11) under the definition of HS norm of tensor products in [33]. In real data analysis with finite number of samples, estimates of HSIC are empirically computed as in equation (12). Here, $K$ is the kernel matrix of $k(x, x')$, $L$ is the kernel matrix of $l(y, y')$, and $H$ is the centering matrix: $\mathbf{I} - \frac{1}{m}\mathbf{1}\mathbf{1}^T$ [33,34]. For this research, normalized version of the empirical HSIC estimate was implemented for computational efficiency and stability under tensorflow. Specific forms of normalized HSIC can be expressed as (13).

$$HSIC(p_{xy}, \mathcal{F}, \mathcal{G}) = \|C_{xy}\|_{HS}^2 = E_{x,y,x',y'}[\langle \phi(x) \otimes \psi(y), \phi(x) \otimes \psi(y)\rangle_{HS}] + \langle \mu_x \otimes \mu_y, \mu_x \otimes \mu_y \rangle_{HS} - 2E_{x,y}[\langle \phi(x) \otimes \psi(y), \mu_x \otimes \mu_y \rangle_{HS}]$$
$$= E_{x,y,x',y'}[k(x,x')l(y,y')] + E_{x,x'}[k(x,x')]E_{y,y'}[l(y,y')] - 2E_{x,y}[E_{x'}[k(x,x')]E_{y'}[l(y,y')]] \quad (11)$$

$$HSIC((X,Y), \mathcal{F}, \mathcal{G}) \equiv \frac{tr(KHLH)}{(n-1)^2}; (H, K, L \in \mathbb{R}^{n \times n})$$
$$(X,Y) \equiv \{(x_1,y_1),(x_2,y_2),\ldots,(x_n,y_n)\}, K_{ij} = k(x_i,y_i), L_{ij} = l(x_i,y_i) \quad (12)$$

$$nHSIC((X,Y), \mathcal{F}, \mathcal{G}) \equiv \frac{\langle \widetilde{K}, \widetilde{L} \rangle_F}{\|\widetilde{K}\|_F \|\widetilde{L}\|_F} = \frac{tr(\widetilde{K}\widetilde{L})}{\sqrt{tr(\widetilde{K}\widetilde{K})tr(\widetilde{L}\widetilde{L})}} \quad (13)$$

Here, kernel matrices: $\widetilde{K}, \widetilde{L}$ are centered matrices of $K$ and $L$ by $H$, which can be defined as $\widetilde{K} = HKH$, $\widetilde{L} = HLH$ [35]. For individual kernels: $k, l$, Radial Basis Function (RBF) kernels (Gaussian Kernels) were assumed for generalized approaches. Based on its characteristics as an alternative measure of MI and its ability to detect non-linear dependencies in a non-parametric framework, HSIC is regarded as a strong quantitative metric to evaluate relevance between variables without conventional high computational costs from MI estimations.

By setting the sub-objective of LightHCG as minimizing the HSIC between variables in sub-space $Z_1$ and glaucoma status(*denoted as Y) and maximizing the HSIC between variables in sub-space $Z_2$ and glaucoma status, glaucoma-relevant information was guided to be extracted to $Z_2$ for causal representation learning, while preventing the deterioration of the original VAE image reconstruction objective by indirectly locating non-glaucoma related retinal information to $Z_1$. Here, to focus on individual latent variables itself, HSIC was first computed feature-wise, and then was incorporated to a mean value according to the type of subspace. Furthermore, to emphasize the importance of $Z_2$, an additional weight parameter: $\omega$ was added to the mean HSIC score from $Z_2$ when defining the min-max objective of (14).

$$\textbf{Objective}: \min_{Z} \frac{1}{J}\sum_{j=1}^{J} nHSIC(Z_{1,j}, Y) - \omega \frac{1}{M}\sum_{m=1}^{M} nHSIC(Z_{2,m}, Y) \quad (14)$$

*(3) LightHCG Loss*

Based on the proposed structure of the LightHCG model, total loss of LightHCG can be decomposed into three partial losses: GAE loss, Convolutional VAE loss, and HSIC loss. Specifically, the regularization term in the Convolutional VAE loss was further weighted ($\beta_1$) to lower the possibility of collapse (convergence to single image) in image reconstruction tasks. Definition of LightHCG loss can be summarized as equation (15).

$$L_{total} = \varphi_1 L_{cvae} + \varphi_2 L_{gae} + \varphi_3 L_{HSIC(1)} + \varphi_4 L_{HSIC(2)}$$
$$L_{cvae} = BCE(\hat{X}, X) + \beta_1 \frac{1}{2}\sum z_{ij}^2$$
$$L_{gae} = \frac{1}{n}\sum_{i=1}^{n} \|Z'^{(i)} - \hat{Z}'^{(i)}\|^2 + \lambda\|A\|_1 + \alpha h(A) + \frac{\rho}{2}|h(A)|^2$$
$$L_{HSIC(1)} = \frac{1}{J}\sum_{j=1}^{J} nHSIC(Z_{1,j}, Y) - \omega \frac{1}{M}\sum_{m=1}^{M} nHSIC(Z_{2,j}, Y)$$
$$L_{HSIC(2)} = \frac{1}{\binom{M}{2}}\sum_{i=1}^{M-1}\sum_{j>i}^{M} nHSIC(Z_{2,i}, Z_{2,j}), \quad Z' = \langle Z_2, Y \rangle \quad (15)$$

Each partial loss within the total loss of LightHCG was given individual weight parameters: $\varphi_1$ to $\varphi_4$. For HSIC loss, apart from the primary information loss introduced in (14), which was defined as $L_{HSIC(1)}$, an additional interaction information loss $L_{HSIC(2)}$ was introduced in order to prevent excessive redundancy between features in causal latent space $Z_2$.

Based on the stage of epochs within training, different weights: $\varphi_1$ to $\varphi_4$, were applied to achieve balance among three objectives: image reconstruction under valid latent space retrieval, glaucoma information disentanglement, and causal representation learning. In early stages, reconstruction loss was most emphasized for valid representations of retinal images. In mediate stages, casual loss and disentanglement loss was most emphasized for latent space optimization. For the final stages of training, reconstruction loss was once again emphasized for the LightHCG to return valid retinal images that correspond to latent states extracted from the original data.

### C. Experimental Design

For LightHCG fitting, cropped retinal fundus images from the ACRIMA dataset and corresponding glaucoma status were implemented as input data. For reparameterization within CVAE settings, standard normal distribution was assumed for randomness. Specific architectures for CVAE and GAE were set as TABLE I, TABLE II. Unlike most advanced CNN models, this research only assumed 4 convolutional2D layers with 16, 32, 16, 16 filters for the CVAE encoder to create a light-weight representation model. Regarding the dimension of latent space Z, $Z_1 \in \mathbb{R}^{4 \times N}$ and $Z_2 \in \mathbb{R}^{3 \times N}$ were assumed. Specifically, number of latent features in $Z_2$ were set based on the number of target factors (optic cupping, neuroretinal rim, neovascularization) dealt in Fig 1. Thus, corresponding input dimension of GAE was set as 4(=3+1), as glaucoma status variable Y was concatenated for weak supervision within causal discovery. To incorporate the weighted adjacency matrix within GAE architectures, an additional adjacency weight layer was attached to the front of the GAE decoder. Unlike general initial weights in neural networks, initial weights for adjacency layer were set as zero to prevent starting point-induced bias within DAG search. Furthermore, within the adjacency weight layer, diagonal elements and links from Y to $Z_2$ were blacklisted for faster convergence.

TABLE I
HYPER PARAMETERS SETTINGS 1

| Type | Network Architecture |
|---|---|
| Encoder | 16 Conv2D 4x4 filters, stride=2, silu<br>32 Conv2D 6x6 filters, stride=3, silu<br>16 Conv2D 4x4 filters, stride=2, silu<br>16 Conv2D 3x3 filters, stride=2, silu<br>128, FC layer(Flattened), ELU<br>16, FC layer, ELU<br>14(=(4+3)*2), FC layer, Linear |
| Decoder | 16, FC layer, ELU<br>128, FC layer(Flattened), ELU<br>8x8x16 FC layer, ELU<br>Reshape(8,8,16) layer<br>16 Conv2DTranspose 3x3 filters, stride=2,silu<br>32 Conv2DTranspose 4x4 filters, stride=2,silu<br>16 Conv2DTranspose 6x6 filters, stride=3,silu<br>3 Conv2DTranspose 4x4 filters, stride=2,sigmoid |

*Silu(Swish): Activation function proposed by [36].

TABLE II
HYPER PARAMETERS SETTINGS 2

| Type | Parameter settings |
|---|---|
| $f_1$(GAE) | 4*4(# of nodes), Sparse Layer, ELU<br>4*4, Sparse Layer, ELU<br>4, Sparse Layer, Linear |
| $f_2$(GAE) | 4, Adjacency Layer, Linear (**attached**)<br>4*4, Sparse Layer, ELU<br>4*4, Sparse Layer, ELU<br>4, Sparse Layer, Linear |

*Sparse Layer: Masked FC Layer based on additive modeling approach

Meanwhile, for general additive modelling within GAE frameworks, binary weight masks were Hadamard multiplied to ensure feature-wise sub-neural network constructions. For each latent feature within GAE, a two-layer NN with 4 nodes(layer-wise) was assigned to extract complex non-linear causal links. In algorithmic aspects, masking FC layers within the GAE were computed by converting binary mask matrices into kernel constraints under tensorflow's keras.constraints.Constraint(). All basic structures and weights within the LightHCG model were based on Python Tensorflow version 2.19.0. Training procedures and experiments based on the proposed LightHCG framework were computed under Google Colab's basic T4 NVIDIA GPU environments.

Within training under full batch gradient descent algorithms, balance between sub-objectives was achieved by implementing different loss functions and weights: $\varphi_1$ to $\varphi_4$ according to the corresponding sub-structure and stages which were dealt in Section **B.(3)** (**Algorithm 1**).

**Algorithm 1** LightHCG Training framework (400 epochs)

-**Initial settings**: i = 0 , maximum epoch(K) = 400;
-Set initial $\alpha = 0.6, \rho = 0.1, \gamma = 0.9, \beta = 1.01$, Adam Optimizer
-**While i < K**:
   1. Compute $L_{cvae}, L_{gae}, L_{HSIC(1)}, L_{HSIC(2)}$; $\omega = 1.5, \beta_1 = 0.001, \lambda = 0$
   2. Compute Total LightHCG loss: $L_{total}$
   **IF i < 50**:
    -Total LightHCG loss = $\boldsymbol{L_{total}}(\varphi_1 = 1, \varphi_2 = 1, \varphi_3 = 0, \varphi_4 = 0)$
   **ELIF i<100**:
    -Total LightHCG loss = $\boldsymbol{L_{total}}(\varphi_1 = 1, \varphi_2 = 1, \varphi_3 = 5, \varphi_4 = 0)$
   **ELSE**:
    -Total LightHCG loss = $\boldsymbol{L_{total}}(\varphi_1 = 2, \varphi_2 = 1, \varphi_3 = 5, \varphi_4 = 0.5)$
   3. Update weights via gradient descent methods:
    1) Update adjacency matrix *A* and GAE weights with $\boldsymbol{L_{gae}}$
     *(learning rate: 0.005, 0.002)
    2) Update remaining CVAE structure weights with $\boldsymbol{L_{total}}$
     *(learning rate: 0.0005)
    3) Update $\alpha, \rho$ using **(8)**
   5. Update i = i+1;
**Return**: weighted adjacency matrix *A*, Latent Space $Z_1, Z_2$

After training the LightHCG model, success in latent causal representation learning regarding glaucoma was checked in both quantitative and qualitative aspects. For quantitative checks, Structured Hamming Distance(SHD) between the extracted causal DAG and ground truth causality(Fig 1.) under best possible match in topological aspects, Mutual Information between features in $Z_2$ and retinal glaucoma status, and GAE loss convergence were evaluated. When computing SHD, extracted weighted adjacency matrix was binarized by converting weights that correspond to top 25% absolute weights to 1 and the rest as 0. For qualitative checks, retinal fundus visualization results from varying individual latent factors in $Z_2$ were compared with features of ground truth physical factors of glaucoma. Specifically, this research randomly sampled 50 retinal fundus images from training dataset and then visualized the obtained mean absolute difference of generated images induced when varying individual each latent causal factor in $Z_2$ (*mean absolute difference computed by comparing the baseline reconstructed image and generated image under latent factor variation). Sequentially checking features of changes in corresponding visualizations, each latent factor in $Z_2$ was evaluated whether it can correspond to neuroretinal rim, optic cup, and neovascularization status.

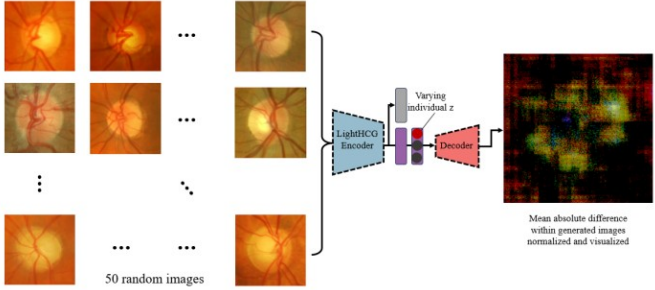

**Fig 3. Visual summary of generating images of mean absolute difference when varying individual latent causal factors**

### D. Glaucoma Classification with LightHCG

After success in latent causal representation retrieval regarding the ground truth glaucoma causal graph dealt in Fig 1., three latent causal features from LightHCG (latent space $Z_2$) were used to fit a downstream task model for automatic glaucoma detection. For the downstream task model, a shallow DNN with 3 fully-connected hidden layers was implemented to further reduce parameter usage within glaucoma classification. Specific model architecture for the LightHCG-based detection model was defined as in TABLE III. When fitting the downstream LightHCG model, identical training dataset used in the previous representation learning phase was also implemented, whereas evaluation of the model was based on the test dataset introduced in Section III.A.

Furthermore, to compare competitiveness and efficiency of LightHCG frameworks with existing approaches, this research also additionally fitted three widely implemented models: InceptionV3, MobileNetV2, and VGG16 under transfer learning approaches (*weights pre-trained with ImageNet). For downstream predictions, additional layers of global average pooling2D and a fully connected hidden layer with 64 nodes and an ELU activation function were attached to the output layer of each model. Under identical training/testing dataset usage, glaucoma classification performance was evaluated with five different metrics: classification accuracy, precision score, recall score, AUC score, and f1-score, which were also implemented in LightHCG evaluations. Total training epochs, learning rate, types of optimizer and batch size were all identically set as 300, 1e-4, Adam and 100.

TABLE III
HYPER PARAMETERS SETTINGS IN DNN

| Type | Network Architecture |
|---|---|
| DNN | 32(# of nodes), FC layer, ELU<br>Batch Normalization( )<br>Dropout(0.05)<br>64, FC layer, ELU<br>Batch Normalization( )<br>32, FC layer, ELU<br>Batch Normalization( )<br>1, FC layer, Sigmoid |

## IV. EXPERIMENTAL RESULTS

### A. Representation Learning Results from LightHCG fitting

Results of glaucoma-related latent causal representation retrieval with LightHCG were deduced as follows. Regarding Total LightHCG loss, convergence was checked after 300 epochs. For specific loss values, $L_{cvae}$=0.5782, $L_{gae}$=0.0705, $L_{HSIC(1)}$= - 0.9861 were deduced under stable convergence (Fig 4). Thus, successful fitting and latent causal space retrieval was checked under retinal fundus image data usage. For information disentanglement checks, MI between variables in $Z_1$ and glaucoma status, and MI between variables in $Z_2$ and glaucoma status were quantified with the mutual_info_regression() function from python's sklean.feature_selection package (*number of neighbors for MI estimation set as 5). With the mean mutual information from $Z_1$ and $Z_2$ each being 0.0109 and 0.4547, it was found plausible to conclude that glaucoma non-related and glaucoma related features were successfully disentangled into subspaces $Z_1$ and $Z_2$ with high significance. In individual aspects, 0.0, 0.0135, 0.0300 and 0.0 were deduced as MI values that correspond with $Z_{1,0}$ to $Z_{1,3}$, whereas 0.6860, 0.0023 and 0.6758 were deduced as MI values that correspond with $Z_{2,0}$ to $Z_{2,2}$.

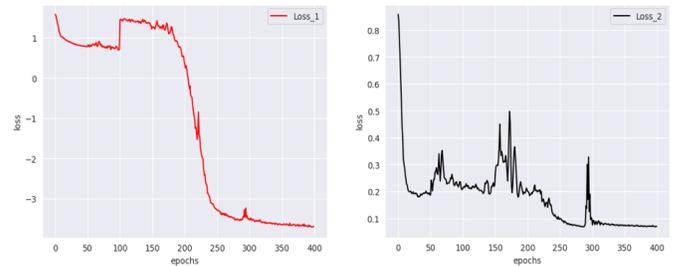

**Fig 4. Visualization of loss values from LightHCG fitting. (Left) $L_{total}$ per epoch (Right) $L_{gae}$ per epoch**

Next, based on success found in LightHCG fitting results and HSIC-based information disentanglement, validity of causal space retrieval was checked by comparing the extracted weighted adjacency matrix with ground truth glaucoma causal DAG. After binarization of the extracted adjacency matrix under the top 25% criterion, latent causal DAG from LightHCG was deduced as in Fig 5. Under best possible match between nodes, SHD between the extracted and ground truth causal DAG was deduced as 2.0, which implies that causal factors in $Z_2$ can be interpreted as valid representations of ground truth physical factors. Furthermore, under topological aspects, it was found that two latent factors: $Z_{2,0}$ and $Z_{2,2}$ may have high possibility of one-to-one corresponding with neuroretinal rim thinning and optic cupping. This strong possibility was further checked under visual disentanglement checks dealt in III.C.

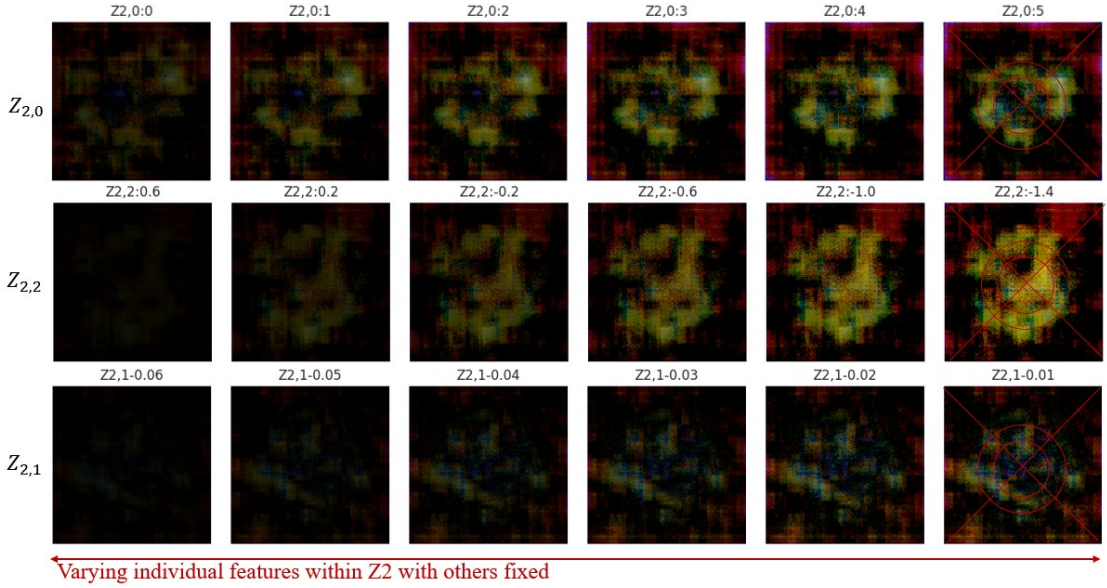

**Fig 7. Visual disentanglement checks by varying latent causal factors in $Z_2$.** Mean absolute difference between baseline reconstructed image values and reconstructed image values from varying individual variables in $Z_2$ were sequentially visualized as above.

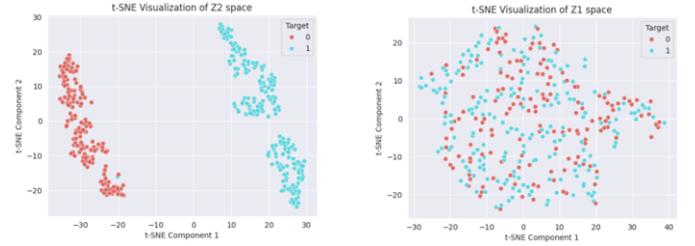

**Fig 6. (Left) t-SNE result ($Z_2$), (Right) t-SNE result ($Z_1$)**

perplexity and learning rate within the TSNE( ) function, 20 and 100 were set as hyper parameters for t-SNE analysis. After fitting, t-SNE results for $Z_1$ and $Z_2$ space were visualized as Fig 6 (target: glaucoma status). As can be seen in Fig 6., use of $Z_2$ exhibited clear separation between glaucoma/normal retinal images within the t-SNE plane, whereas use of $Z_1$ exhibited extreme entanglements. Thus, it was found that HSIC based disentanglement successfully allocated glaucoma non-related/ related information to $Z_1$ and $Z_2$, which leads to robust representations of glaucoma detection in LightHCG.

### B. Visual Causal disentanglement checks

With the success achieved in retrieving valid latent causalities and disentangling glaucoma related information in LightHCG training, qualitative checks regarding features in $Z_2$ were carried out for further validation of the possibility of one-to-one correspondence. Visualizations of mean absolute difference between reconstructed images based on varying individual features in $Z_2$ were computed as in Fig 7. Here, all mean absolute differences between reconstructed retinal images were normalized (row-wise) to values between 0 and 1. Additionally, generated difference values which are under the top 25% criterion were clipped to 0 to focus only on consistent changes. (*Interval definitions for variations of individual $Z_{2,i}$

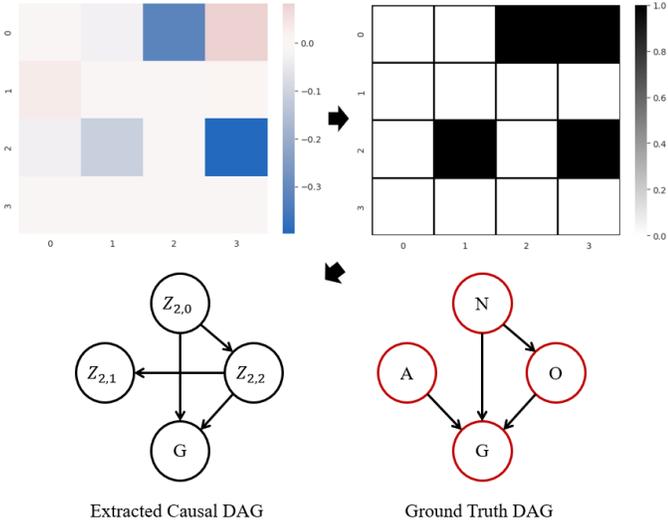

**Fig 5. (Upper Left) Extracted Latent adjacency matrix, (Upper Right) Binarized $A$, (Lower Left) Extracted Causal DAG, (Lower Right) Ground Truth DAG**

Apart from causal retrieval checks, HSIC disentanglement success was again checked under visual aspects using t-SNE algorithms from python's sklearn.manifold package. For were based on the min, max values of each variable. Baseline $Z_{2,i}$ values for sequential comparison were each set as -2, 0.8, and -0.07).

For variables: $Z_{2,0}$, features and location which directly correspond to the general characteristics of neuroretinal rim change were found to be affected when varying individual $Z_{2,0}$ values (Fig 7). As $Z_{2,0}$ values were increased, a clear ring-shaped variation around the optic cup region was found to be generated in common within 50 random retinal images. Considering the fact that rim thinning around the optic cup is a representative characteristic of neuroretinal rim change in glaucoma patients, and considering previous success in latent causal DAG retrieval, it was found plausible to conclude that the extracted $Z_{2,0}$ has high possibility of having one-to-one correspondence with the neuroretinal rim.

Meanwhile, when varying $Z_{2,2}$, variations within the optic cup area, which is located in the center of retinal fundus images, were found to be affected. Though parts of the neuroretinal rim area were also found to be altered, considering the fact that optic cup enlargement due to glaucoma mainly accompanies vertical changes in the original rim area(inferior, superior area) and the fact that mean absolute difference was most vivid in the center of the original optic cup area, it was found that $Z_{2,2}$ can be interpreted as a valid one-to-one corresponding representation of optic cup features in the glaucomatous retina.

In contrast, regarding latent factors $Z_{2,1}$, no significant characteristics that can correspond with neovascularization or abnormality in retinal vessels were found to be visible, which leads to the fact that although $Z_{2,1}$ was incorporated into the extracted latent causality, $Z_{2,1}$ cannot be considered as a valid representation of significant physical factors of glaucoma which were dealt in Section II. Fig 1.

Thus, through the LightHCG framework, it was found that not only latent causal identification, but also latent causal representation retrieval that can one-to-one correspond with neuroretinal rim and optic cup status can be successfully achieved. This implies that LightHCG can not only return robust glaucomatous representations for downstream tasks, but can also enable valid causal simulations regarding glaucoma treatments or risk factors. By returning visualization of changes generated when variables $Z_{2,0}$ and $Z_{2,2}$ are intervened through the fitted GAE, LightHCG can aid clinicians to simulate or quantify the effect of certain factors on alleviating the risk of glaucoma under latent causality consideration. Returning predictions or representations under partial comprehension regarding physical structures within the glaucoma mechanism, LightHCG can help current AI-driven glaucoma detections to overcome limitations as a black box which only focuses on correlation structures within data.

*C. Glaucoma Detection under LightHCG*

Based on success in LightHCG model fitting, enhancements in automatic glaucomatous retinal fundus classification under LightHCG encodings were checked for clinical applications. Specifically, only three latent causal representations from $Z_2$ were used to create a robust, lightweight downstream model.

Here, to check competitiveness and advantages of using LightHCG over existing advanced vision model approaches, performance from MobileNetV2, InceptionV3 and VGG16 based transfer learning models and their parameter space size were compared with the LightHCG approach. Classification results from processing test data were deduced as TABLE IV, TABLE V and Fig 8.

Within metrics, the LightHCG-based model achieved the highest accuracy, precision score and f1-score among all models. Specifically, the LightHCG-based model exhibited 92.63% accuracy score, 91.37% recall score, 95.24% precision score, 93.26% f1-score and 97.13% AUC score in test data classification tasks. When compared to other models, LightHCG achieved a 5.66%p higher accuracy score than the InceptionV3-based model and a 8.57%p higher precision score than the VGG16-based model. Among non-LightHCG based models, MobileNetV2 based model succeeded in achieving higher AUC scores than the LightHCG approach. However, considering superiority of the LightHCG-based model in three metrics(acc, precision, f1-score) and the 3.16%p precision score gap between the two models, LightHCG model was found to achieve higher performance than all other models in general terms. In terms of size of parameter space, LightHCG only used approximately 7% of MobileNetV2, 0.8% of InceptionV3 and 1.1% of VGG16 based models, which implies that LightHCG achieved superior glaucoma detection performance while using 93%~99% less parameters than existing advanced vision models. Thus it was found that with appropriate HSIC-based disentanglement and latent causal representation learning from retinal fundus images, high performing downstream task models for glaucoma detection can also be achieved with extremely less use of parameters, which makes the LightHCG framework suitable for both reliable glaucoma diagnosis and intervention or treatment simulations in clinical aspects.

TABLE IV
GLAUCOMA CLASSIFICATION PERFORMANCE ON TEST DATA

| Type | Acc | Recall | Precision | F1 score | AUC score |
|---|---|---|---|---|---|
| MobileNetV2 | 92.35% | **94.42%** | 92.08% | 93.23% | **98.06%** |
| InceptionV3 | 86.97% | 87.82% | 88.72% | 88.27% | 93.96% |
| VGG16 | 87.82% | 92.39% | 86.67% | 89.43% | 94.54% |
| **LightHCG** | **92.63%** | 91.37% | **95.24%** | **93.26%** | 97.13% |

*Highest metric value emphasized in bold*

TABLE V
PARAMETER SPACE SIZE COMPARISON

| Type | Total Number of Parameters |
|---|---|
| MobileNetV2 Based | 2,504,133 |
| InceptionV3 Based | 22,196,389 |
| VGG16 Based | 14,813,381 |
| **LightHCG ($Z_2$ Encoder+DNN)** | **168,155** |

*Parameter count contains weights from additional downstream layers. Transfer learning based on (include_top=False) settings.*

Meanwhile, for a broader and more extensive evaluation of LightHCG's efficiency and performance in glaucoma detection tasks, AUC score from LightHCG was compared to previous works that incorporate deep learning architectures other than advance vision models dealt in TABLE IV. AUC scores of models from previous works were summarized in TABLE VI.

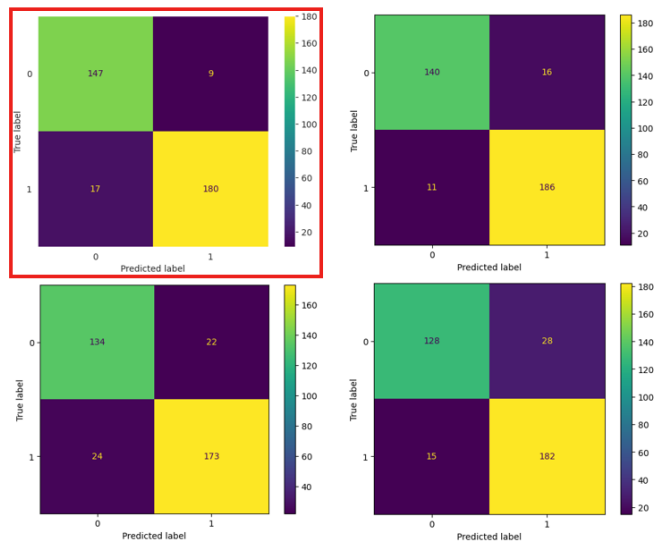

**Fig 8. Confusion Matrix from Glaucoma Classification Tasks. Upper Left(Red Box):** LightHCG, Upper Right: MobileNetV2, Lower Left: InceptionV3, Lower Right: VGG16

TABLE VI
AUC SCORE COMPARISON BETWEEN EXISTING WORKS

| Type | Dataset | AUC (test data) |
| --- | --- | --- |
| ViT [7] | Merged data (ORIGA, REFUGE, ODIR, HRF,ORIGA, DRISHTI) | 92.60% |
| CrossViT [7] | Merged data (ORIGA, REFUGE, ODIR, HRF,ORIGA, DRISHTI) | 96.87% |
| EfficientNet-B5[37] | ODIR-5K | 94.53%(off-site) |
| EfficientNet-B5+ MLD+SAM[37] | ODIR-5K | 95.02%(off-site) |
| Hybrid Model (CNN+SVM)[9] | ACRIMA | 90.60% |
| Hybrid Model (CNN+AdaBoost)[9] | ACRIMA | 92.80% |
| 3D ResNet [10] | Spectral-domain OCT volume data | 96.90% |
| *LightHCG (Proposed) | ACRIMA | **97.13%** |

*Highest metric value emphasized in bold*

Results show that although the proposed LightHCG-based downstream model was fitted under an extreme train : test ratio of 50:50, and was constructed with significantly less parameters than advanced vision models, LightHCG exhibits substantial superiority to a variety of models introduced in prior studies. Thus, it was found that the proposed LightHCG framework can not only enable visual simulations or intervention analysis regarding clinical treatments of glaucoma, but can also significantly contribute in enhancing the performance and efficiency of current AI-based glaucoma diagnosis models by providing robust latent causal representations under far fewer parameter usage.

## V. CONCLUSION

In this work, a causality-based reliable, yet extremely lightweight AI-driven glaucoma diagnosis model: LightHCG was constructed to enhance the performance of current vision AI-based glaucoma detection approaches and to increase the possibility of deep learning-based clinical intervention analysis regarding glaucoma treatments. Specifically, LightHCG was constructed under a Convolutional VAE(CVAE)-based latent representation extraction framework and a two stage disentanglement framework, which first incorporates a Hilbert Schmidt Independence Criterion(HSIC)-based disentanglement in latent space fitting, and then applies a GAE-based causal representation learning framework in optimizing a pre-defined causal latent sub-space. Within the Convolutional VAE architecture, primary latent space Z was split into two latent sub-spaces: $Z_1$ and $Z_2$, which were each assumed to incorporate glaucoma non-related and related features, sequentially. Additionally, $Z_2$ was defined as a causal sub-space that represents core physical factors found within the ground truth glaucoma causal mechanism (Fig 1). Successfully separating relevant information to $Z_1$ and $Z_2$ with HSIC and extracting the underlying latent causal DAG with Graph Autoencoder(GAE)-based causal structure discovery algorithms, robust causal representations that can one-to-one correspond with neuroretinal rim status and optic cupping were achieved from only retinal fundus images and corresponding labels within the ACRIMA dataset.

Based on successful latent causal space retrieval, performance of LightHCG in glaucoma diagnosis tasks were evaluated by comparing a LightHCG encoder + shallow DNN based model with advanced vision models, such as transfer learning-based InceptionV3, VGG16 and MobileNetV2. Results show that the LightHCG-based downstream model achieved the highest classification accuracy, precision score and f1-scores among all candidates despite using 93%~99% less parameters than other models (TABLE IV). Furthermore, LightHCG-based glaucoma detection model also exhibited superior performance when compared to previous representative works with an AUC score of 97.13% (TABLE VI). Thus, it was found that the proposed LightHCG framework can not only enable visual simulations or intervention analysis regarding clinical treatments of glaucoma by providing valid latent causal links, but can also significantly contribute in enhancing the performance and efficiency of current AI-based glaucoma diagnosis models with only three causal representations under far fewer parameter usage. Moreover, the LightHCG framework can enable vision models to comprehend the underlying physical mechanism of glaucoma, which, in turn, can lead to more robust estimations compared to existing works, effectively aiding glaucoma diagnosis in clinical fields.

However, several limitations were also visible throughout the validation process regarding LightHCG frameworks. First, due to the absence of meta data which can correspond to physical causal factors of glaucoma, quantitative disentanglement checks regarding individual latent causal representations from LightHCG were highly limited, leading to a reliance on qualitative disentanglement checks. Second, though the LightHCG framework succeeded in retrieving valid causal representations that can one-to-one correspond to neuroretinal rim status and optic cupping, representations that can correspond to neovascularization or abnormality in retinal vessels were not achieved from latent causal DAG learning, leading to partial comprehensions regarding the overall human glaucoma mechanism. Presuming that these limitations are being dealt, LightHCG is believed to not only enable accurate intervention analysis in both visual and quantitative aspects, but also further enhance the concept of physical mechanism comprehension within computer vision AIs for glaucoma diagnosis.

VI. REFERENCE

[1] C.X, Hu et al., "What Do Patients With Glaucoma See? Visual Symptoms Reported by Patients With Glaucoma", Am J Med Sci, Vol. 348, no. 5, pp. 403-409, 2014.
[2] National Eye Institute, "Glaucoma", Eye Conditions and Diseases, 2025, https://www.nei.nih.gov/learn-about-eye-health/eye-conditions-and-diseases/glaucoma.
[3] S, Shan et al., "Global incidence and risk factors for glaucoma: A systematic review and meta-analysis of prospective studies", Journal of Global Health, Vol. 14, pp. 04252, 2024.
[4] J, Camara et al., "A Comprehensive Review of Methods and Equipment for Aiding Automatic Glaucoma Tracking", Diagnostics(Basel), Vol. 12, no. 4, pp. 935, 2022.
[5] J, Phu et al., "Standard automated perimetry for glaucoma and diseases of the retina and visual pathways: Current and future perspectives", Progress in Retinal and Eye Research, Vol. 104, 101307, 2025.
[6] N, Akter et al., "Glaucoma detection and staging from visual field images using machine learning techniques", PloS One, Vol. 20, no. 1, e0316919, 2025.
[7] M, Wassel et al., "Vision Transformers Based Classification for Glaucomatous Eye Condition", In: 2022 26th International Conference on Pattern Recognition (ICPR), pp. 5082-5088, 2022.
[8] S.M, Saqib et al., "Cataract and glaucoma detection based on Transfer Learning using MobileNet", Heliyon, Vol. 10, no. 17, e36759, 2024.
[9] C, Oguz et al., "A CNN-based hybrid model to detect glaucoma disease", Multimedia Tools and Applications, Vol. 83, pp. 17921-17939, 2024.
[10] A.R, Ran et al., "Detection of glaucomatous optic neuropathy with spectral-domain optical coherence tomography: a retrospective training and validation deep-learning analysis", The Lancet Digital Health, Vol. 1, no. 4, pp. e172-e182, 2019.
[11] Y, George et al., "Attention-guided 3D-CNN Framework for Glaucoma Detection and Structural-Functional Association using Volumetric Images", IEEE J Biomed Health Inform, Vol. 24, no. 12, pp. 3421-3430, 2020.
[12] W, Zhou et al., "EARDS: EfficientNet and attention-based residual depth-wise separable convolution for joint OD and OC segmentation", frontiers in Neuroscience, Vol. 17, 1139181, 2023.
[13] National Eye Institute, "Glaucoma and Eye Pressure", Glaucoma, 2024, https://www.nei.nih.gov/learn-about-eye-health/eye-conditions-and-diseases/glaucoma/glaucoma-and-eye-pressure.
[14] S.G, Asrani et al., "The relationship between intraocular pressure and glaucoma: An evolving concept", Vol. 103, 101303, 2024.
[15] A, McCalla et al., "Association Between Cup-to-Disc Ratio and Structural and Functional Damage Parameter in Glaucoma: Insights From Multiparametric Modeling", Translational Vision Science & Technology, Vol. 14, no. 4, pp. 17, 2025.
[16] F.A, Medeiros et al., "Evaluation of Progressive Neuroretinal Rim Loss as a Surrogate Endpoint for Development of Visual Field Loss in Glaucoma", Ophthalmology, Vol. 121, no. 1, 2013.
[17] P.J, Airaksinen et al., "Neuroretinal Rim Area in Early Glaucoma", American Journal of Ophthalmology, Vol. 99, no. 1, pp. 1-4, 1985.
[18] R.N, Weinreb et al., "The Pathophysiology and Treatment of Glaucoma", JAMA, Vol. 311, no. 18, pp. 1901-1911, 2015.
[19] A.J, Tatham et al., "The Relationship Between Cup-to-Disc Ratio and Estimated Number of Retinal Ganglion Cells", Investigative ophthalmology & visual science, Vol. 54, no. 5, pp. 3205-3214, 2013.
[20] A, Katsimpris et al., "Intraocular pressure, primary open-angle glaucoma and the risk of retinal vein occlusion: A Mendelian randomization mediation analysis", Eye, Vol. 38, pp. 3347-3351, 2024.
[21] C, Mishra et al., "Neovascular Glaucoma", StatPearls, 2022, https://www.ncbi.nlm.nih.gov/books/NBK576393/.
[22] https://figshare.com/articles/dataset/CNNs_for_Automatic_Glaucoma_Assessment_using_Fundus_Images_An_Extensive_Validation/7613135
[23] A.D, Pinto et al., "CNNs for automatic glaucoma assessment using fundus images: an extensive validation", BioMedical Engineering OnLine, Vol. 18, no. 29, 2019.
[24] Kaggle, https://www.kaggle.com/datasets/orvile/acrima-glaucoma-assessment-using-fundus-images
[25] D.P, Kingma and M, Welling, "Auto-Encoding Variational Bayes", arXiv:1312.6114v11, pp. 1-14, 2022.
[26] I, Ng, S, Zhu, Z, Chen and Z, Fang, "A graph autoencoder approach to causal structure learning", NeurIPS 2019 Workshop, pp. 1-8, 2019.
[27] X. Zheng et al., "Dags with no tears: Continuous optimization for structure learning", Advances in neural information processing systems, arXiv:1803.01422v2, pp. 1-22, 2018.
[28] Y. Yu et al., "DAG-GNN: DAG Structure Learning with Graph Neural Networks", arXiv:1904.10098v1, pp. 1-12, 2019.
[29] J, Pearl, "Causal inference in statistics: An overview. Statistics Surveys", Vol. 3, pp. 96-146, 2009.
[30] J, Pearl, "Comment: Graphical models, causality and intervention", Statistical Science, Vol. 8, no. 3, pp. 266-269, 1993.
[31] C, Li et al., "An efficient augmented Lagrangian method with applications to total variation minimization", Computational Optimization and Applications, Vol. 56, pp. 507-530, 2013.
[32] W.D.K, Ma et al., "The HSIC Bottleneck: Deep Learning without Back-Propagation", arXiv:1908.01580v3, pp. 1-9, 2019.
[33] A, Gretton et al., "Measuring statistical dependence with Hilbert Schmidt norms", In: International conference on algorithmic learning theory, Berlin, Heidelberg: Springer Berlin Heidelberg, pp. 63-77, 2005.
[34] K, Sakamoto and I, Sato, "End-to-End Training Induces Information Bottleneck through Layer-Role Differentiation: A Comparative Analysis with Layer-wise Training", arXiv:2402.09050v2, pp. 1-40.
[35] T, Gorecki et al., "Independence test and canonical correlation analysis based on the alignment between kernel matrices for multivariate functional data", Artifical Intelligence Review, Vol. 53, pp. 475-499, 2020.
[36] P, Ramachandran et al., "Searching for Activation Functions", arXiv:1710.05941v2, pp. 1-13, 2017.
[37] O, Sivaz and M, Aykut, "Combining EfficientNet with ML-Decoder classification head for multi-label retinal disease classification", Neural Computing and Applications, Vol. 36, pp. 14251-14261, 2024.